\documentclass[10pt]{article}
\usepackage[T1]{fontenc}
\usepackage[utf8]{inputenc}
\usepackage{authblk}
\usepackage{arxiv}
\usepackage{graphicx}

\usepackage{amsmath}
\usepackage{amssymb}
\usepackage{amsfonts}

\newcommand{\bs}[1]{\boldsymbol{#1}}

\newcommand{\dpar}[2]{\frac{\partial #1}{\partial #2}}

\title{Thermodynamics of learning physical phenomena}

\author[1]{El\'ias Cueto}
\author[2,3]{Francisco Chinesta}

\affil[1]{{\small ESI Group chair. Aragon Institute of Engineering Research (I3A). Universidad de Zaragoza. Zaragoza, Spain.}}
\affil[2]{{\small ESI Group chair. PIMM Lab. ENSAM Institute of Technology. Paris, France.}}
\affil[3]{{\small CNRS@CREATE, 1 Create Way, \#08-01 Create Tower, Singapore 138602.}}

\begin{document}

\maketitle

\begin{abstract}
Thermodynamics could be seen as an expression of physics at a high epistemic level. As such, its potential as an inductive bias to help machine learning procedures attain accurate and credible predictions has been recently realized in many fields. We review how thermodynamics provides helpful insights in the learning process. At the same time, we study the influence of aspects such as the scale at which a given phenomenon is to be described, the choice of relevant variables for this description or the different techniques available for the learning process.
\end{abstract}

\section{Introduction}

In a 2009 compilation of essays, the fourth paradigm of science was first described \cite{fourth-paradigm}. After centuries of science based on observation---the empirical period of the first paradigm---came a period based on the establishment of scientific laws---the second paradigm, think of Newton---and much more recently a period in which simulation took an important role---the third paradigm. Very recently, the authors of this essay argue that we have entered a period in which data plays a prominent role in scientific discovery and where theory and experiments, symbiotically, help data to achieve higher goals. Scientific Machine Learning is precisely a new field in which data coming from scientific experiments is used massively to unveil new, still unknown scientific laws. Some authors have begun to think about an even more recent fifth paradigm of science, in which data is obtained not from experiments, but from simulations \cite{5thparadigm}. This approach is helping scientists to look for the origin of dark matter \cite{nishimichi2019dark} or the structure of protein folding \cite{jumper2021highly}, to name but two of the most relevant examples. Other data-driven approaches do not look for a closed-form scientific expression describing a particular phenomenon, but substitute phenomenological laws of low epistemic value (typically, constitutive or closure laws) by data \cite{kirchdoerfer2016data,conti2020data,conti2018data,ibanez2017data,ibanez2018manifold,carrara2022model}. {  Of course, this approach is somewhat more limited in terms of extrapolation capabilities. Although it largely exceeds the scope of this paper, the interested reader could find excellent reviews on machine learning in \cite{AMSES-reviewML,mahesh2020machine,mitchell1990machine}, for instance.}

No doubt that this data-intensive approach is revolutionizing science. But, on a much more applied context, scientific machine learning is also revolutionizing industry. In the same way that these techniques can look for general laws of physics, they can also look for rigorous descriptions of the functioning of technical apparatus, thus helping us to develop digital twins in a very efficient manner \cite{chinesta2020virtual,sancarlos2021rom,moya2020digital}.

However, distilling scientific laws about any physical phenomena, regardless of its practical importance, has profound implications. Galileo, for instance, in his book on two new sciences---one of which was mechanics of materials---failed completely in the description of beam bending phenomena. He assumed that rectangular sections of a beam rotate due to bending around an axis passing through the lower surface of the beam, and not through the center of mass, as it is well known today \cite{galilei1914two}. However, he was intelligent enough to think of rotating {\em sections} of the beam, which is actually the cornerstone of the celebrated Euler-Bernoulli-Navier bending theory. Thus, even if his theory was not entirely true, his choice of variables was correct. Or at least it was at that macroscopic scale of description. Of course, at that time the atomic structure of matter was not known, but this extremely fine scale of description is equally valid, albeit much less efficient. Or that of the theory of elasticity within a continuum mechanics framework. Or that of the Cosserat brothers, that includes rotations of the material point---the beam bending theory could be considered itself as a Cosserat theory on a one-dimensional continuum \cite{cosserat1909theorie}. Any of these scales is in principle valid, yet Galileo chose the right one, as did Euler, Bernoulli and Navier. Here, by ``right'' we mean the most useful one in terms of engineering practice. In Section \ref{Statistical} we briefly review the statistical mechanics aspects related to the choice of an appropriate level of description for a given physical phenomenon.

But the employ of machine learning is by no means restricted to unveil unknown physical laws. Still today, and despite the impressive interest on big data approaches in social sciences, the biggest supercomputers in the world continue to be devoted to the simulation of complex phenomena \cite{supercomputers}. Full-field and high-resolution simulations continues to be challenging, and the possibility of developing a promising family of {\em learned simulators} is appealing \cite{stachenfeld2021learned,allen2022physical,klimesch2022simulating}. These learned simulators share the characteristics of being considerably faster than traditional simulation techniques such as finite elements or finite volumes (once trained, of course, this does not take into account the considerable effort invested in training these networks). While traditional techniques employ a non-negligible effort to construct each model, learned simulators, in general, employ reusable architectures that can be employed for different, but related, problems. Additionally, the results of these learned simulators are, in general, as accurate as the data employed to train the network, and do not depend on models whose validity could be compromised by extreme parameter values, for instance. Another advantage includes the possibility of employ these simulators for inverse problems and optimization procedures, since they are based on differentiable networks \cite{um2020solver,coros2021differentiable,heiden2021neuralsim,schenck2018spnets}.

Despite of the above-mentioned advantages, neural networks are not very popular among the scientific community in general, and particularly in the computational mechanics one \cite{zhang2006avoiding,eitel2021promises}. Common pitfalls include large deviations from the expected result arising as a consequence of small perturbations in data, or the well-known phenomenon of overfitting the data (basically, learning the noise in the data), for instance. This has motivated a strong effort of research towards the inclusion of previous knowledge about the physics taking place in the learning procedure. This knowledge could be included in the form of inductive biases, for instance \cite{battaglia2018relational} or the particular form of the differential equation governing the physics.

Recently, very popular methods based on neural networks have been developed for the solution of Partial Differential Equations (PDEs) from data \cite{raissi2019physics} that try to overcome these limitations. These methods, coined globally as Physics-informed Neural Networks (PINNs), allow one to solve PDEs from data by following a scheme that resembles a collocation method in some sense. Thus, if the PDE governing the problem is known---hence the ``physics-informed'' character---, its particular form can be unveiled from the available data. We devote Section \ref{PINNs} to this family of techniques.

There is still another family of machine learning techniques that consider the learning process as a sort of regression procedure operating on an Ordinary Differential Equation (ODE) governing the dynamics of the system. These  techniques can be seen as a particular instance of the proposal by Weinan E \cite{E2017}. These techniques operate when the particular form of the PDE governing the physics is not known, nor sought. Instead, as will be seen, this particular form of seeing the learning problem leverages the vast corps of knowledge around dynamical systems to enforce the right structure of the system (conserved quantities, symmetries, etc.). To this family of techniques we devote Section \ref{dynamical}.

As will be seen throughout this paper, the employ of neural networks for learning physical phenomena is an active field of research, that has produced hundreds of papers in a very limited period of time. R { Recently, methods have been developed that are able to unveil constitutive laws from displacement data alone ---with global force but, notably, no stress data--- which constitute a great breakthrough in this discipline, see \cite{thakolkaran2022nn,flaschel2023automated,marino2023automated}}. Nevertheless, the topic is still in its infancy, and much more research is expected before we will be able to fully understand and predict the behavior of machine learning techniques, particularly neural networks, in this amazing discipline. The paper ends with some conclusions about these and other considerations in Section \ref{conclusions}.

\section{Statistical mechanics of coarse graining}\label{Statistical}

The simplest approach to learn physics, at least conceptually, consists in going down to the molecular dynamics scale, where Newton laws apply, label every molecule and follow them across their travel. This is not a useful approach, obviously, but conceptually is the simplest one. At this scale, everything is reversible. So if we denote by $\bs z=\{\bs q_1, \ldots, \bs q_N, \bs p_1, \ldots, \bs p_N\}$, where $N$ is the number of molecules (on the order of $10^{23}$ to have meaningful results) and $\bs q_i$, $\bs p_i$, $i=1, \ldots, N$, represent the position and momentum of every molecule, the evolution in time of the system is well-know to obey a Hamiltonian evolution, i.e.,
{ 
\begin{equation}\label{Newton}
 \dot{\bs z} = \frac{\partial \bs z}{\partial t} =  \bs L\nabla \mathcal H = \bs L\nabla E,
\end{equation}}
\noindent with $\bs L$ the so-called symplectic (skew-symmetric) matrix and $\mathcal H$ the Hamiltonian of the system, the total energy, $E$. So, learning the system at this scale means learning the particular form of $\bs L$ and $E$ by regression (a form of supervised learning). Distilling these from data will ensure that predictions obtained through Eq. (\ref{Newton}) will conserve energy, thanks to the symplectic structure of the learned description.

Since this approach is by no means practical, we will need to coarse-grain the description of the system. In other words, we will need to represent it with less degrees of freedom than it actually has. This process is described with particular elegance in a paper by Pep Espa\~nol, whose title we borrow for this section \cite{Espanol}. { Alternative approaches can be found, for instance, in \cite{Turkington_2016}.}In essence, the process consists in eliminating those degrees of freedom whose evolution in time is faster, keeping those whose evolution takes place over longer periods of time, thus allowing us to describe the system over longer intervals with less computational effort. At the other side of the spectrum, (equilibrium) thermodynamics is a description of the system that takes into account only invariants, magnitudes that do not evolve in time (typically, mass, momentum, energy).

Each possible level of description involves a set of variables, which we will denote by $\bs z$, regardless of the particular scale, if there is no risk of confusion. Coarser levels of description will provide us with less information, but will involve smaller sets $\bs z$. The molecular dynamics scale given by Eq. (\ref{Newton}) has a typical time scale on the order of the collision time (again, unpractical for our purposes). In any case, every description will be governed by some equation that, very much like Eq.~(\ref{Newton}) predicts the evolution of the variables governing the system. It is also important to realize that, if a clear separation of scales exists between conserved and eliminated degrees of freedom, the resulting description will be Markovian. In other words, the coarse-grained model will not depend on history, but only in the present value of its variables.

{  Even in the case in which the description is Markovian, it is worth noting that many microscopic states could lead to the same macroscopic state. This produces uncertainty in the future evolution of the coarse-grained description of the system, see \cite{Espanol,WE2,gonzalez2021learning}. This uncertainty is equivalent to fluctuation in the evolution, and fluctuation is equivalent to dissipation by the celebrated fluctuation-dissipation theorem \cite{weber1956fluctuation,kubo1966fluctuation}.}

Espa\~nol takes a system with two clearly separated scales to elaborate around this theory: a colloidal suspension \cite{Espanol}. The finest description is obtained, as discussed before, by taking position and momenta of both coloidal particles, $\bs Q_i, \bs P_i$ and solvent molecules, $\bs q_j, \bs p_j$, with $i=1, \ldots, N_\text{col}$, $j=1, \ldots, N_\text{sol}$ the number of coloidal particles and solvent molecules, respectively. At this scale, the energy of the system is composed by the potential and kinetic energy of the particles, and the typical time scale is on the order of picoseconds.

If we are not interest in describing the movement of every solvent molecule, we can substitute them by a continuous hydrodynamic field, in which extensive magnitudes as mass density, momentum density and energy density fields substitute them. In this framework we loose information about the precise location of each molecule, but we have instead information about how many of them are located within a given region of the fluid. Particles entering and leaving the considered region in the fluid produce the fluctuations mentioned before \cite{landau1992hydrodynamic}. These fluctuations are in turn responsible of the Brownian motion of the coloidal particles. Dissipative Particle Dynamics, for instance, could be employed for the description of the solvent at this scale \cite{espanol1995statistical,warren1998dissipative,espanol2017perspective}.

If the coloidal particles are massive, if compared to those of the solvent, their change in position and momentum will be slow, compared to the scale of hydrodynamic interactions. Therefore, one could envisage the elimination of the hydrodynamic field in order to keep position and momenta of the coloidal particles as the sole variables of our system, $\bs z = \{\bs P_i, \bs Q_i\}$, $i=1, \ldots, N_\text{col}$. This description gives rise to a Fokker-Planck (FP) equation. The general FP equation works with a probability density function $\psi(\bs z,t)$ that reflects the probability of finding a particle at a given position at a given time instant. Its evolution in time is given by
\begin{equation}\label{FP}
\frac{\partial \psi(\bs z,t)}{\partial t} = -\frac{\partial}{\partial \bs z}\cdot \left(\bs A(\bs z,t)\psi(\bs z,t)\right) + \frac{1}{2}\frac{\partial}{\partial \bs z}\frac{\partial}{\partial \bs z}:\left( \bs D(\bs z,t)\psi(\bs z,t) \right).
\end{equation}
$\bs A$ is a term that takes into account deterministc effects related to the macroscopic velocity drift, while $\bs D$ is a diffusion tensor related to Brownian effects \cite{keunings2000advances,keunings2004micro}. The FP equation has an equivalent It\^o stochastic differential equation of the form
\begin{equation}\label{Ito}
d\bs z = \bs A(\bs z,t)dt + \bs B(\bs z,t)\cdot d\bs W,
\end{equation}
where $\bs D= \bs B\cdot \bs B^\top$ and $\bs W$ is a multidimensional Wiener process. In the case of our coloidal suspension, the influence of the hydrodynamic field is taken into account by introducing a friction tensor $\bs \zeta_{ij}$ that depends on the relative position of coloidal particles $i$ and $j$. Its equivalent It\^o stochastic differential equation has the form
\begin{equation*}
d\bs Q_i=\frac{\bs P_i}{M_i}dt,\;\; d\bs P_i = \bs F_i^{CC}dt -\sum_j \bs \zeta_{ij}\cdot \bs V_jdt + d\tilde{\bs F}_i,
\end{equation*}
with $\bs V_i = \bs P_i/M_i$, $F_i^{CC}$ a force of interaction among coloidal particles and $\tilde{\bs F}_i$ a stochastic force described by a Wiener process that takes the form $d\tilde{\bs F}_id\tilde{\bs F}_j= 2k_BT\bs \zeta_{ij}dt$ \cite{owens2002computational}.

Still a coarser description can be obtained if the coloidal particles are placed at distant locations from each other. In that case, we can assume that $\bs \zeta_{ij} = \delta_{ij} \bs I \zeta$, where $\zeta$ is a friction coefficient. This gives rise to the set of so-called Langevin equations. All this works well only if the density of coloidal particles is much bigger than that of the solvent. Otherwise, we can not isolate position and momenta of the coloidal particles as the sole variables in our model.

Coarse-graining this model ever further, one could think that the position of the coloidal particles changes much more slowly than momentum, and try to keep $\bs Q_i$ alone as governing variables of our model. This gives rise to the so-called Smoluchowsky equation \cite{wilemski1976derivation}.  One can even think that we are not interested in knowing the position of each and every coloidal particle in our suspension, but on their density over a particular region. This is possible by introducing a concentration field as the sole variable of the system. This gives rise, in turn, to the well-known Fick equation.

All this process ends by considering that we are only interested in the system at equilibrium. The description that takes into account invariants of the system only is the scale of thermodynamics. For such a simple system we have mentioned six different possible scales at which we can describe its dynamics. As mentioned before, at each of these scales a given degree of uncertainty will appear, as a consequence of our lack of knowledge about the precise microscopic state (at the molecular dynamics scale). This uncertainty appears as stochasticity. In general, at each level of description, the least biased distribution is the one that maximizes the entropy functional at that scale.

It is therefore evident that by performing experiments at a given scale---something that depends usually on the available instruments and not on our choice---we are choosing a particular description of the phenomenon under scrutiny that may heavily influence the way in which the system shows to us. In the absence of previous knowledge on the system, we are not even in the position of ensuring the non-Markovian character of the description. Recently, however, data analysis methods have been devised that help in determining the precise number of internal, phenomenological variables for an accurate description of the history of the system \cite{gonzalez2021learning}.

Under this prism, we could ask ourselves if a suitable description of our system is already available in the literature, or if the phenomenon is completely unknown to us. In the former case, this description will be given frequently in the form of a partial differential equation. The general form of this PDE could be known, but not the precise values of its coefficients nor the boundary conditions applying under the laboratory conditions. Should this be the case, the formalism of physics-informed neural networks (PINNs) becomes the natural way to tackle the problem, see Section \ref{PINNs}. If, on the other hand, we do not have any information on the particular form of the law being sought---think of the equations governing dark matter, for instance---a supervised learning approach based on the dynamical systems equivalence should be preferred. These will be deeply analyzed in Section \ref{dynamical} below.

\section{Physics-informed neural networks}\label{PINNs}

Physics-informed neural networks (PINNs) have constituted a great success in computational mechanics and mathematics and are, perhaps, the responsible of a very active research activity in the field \cite{raissi2019physics,mao2020physics,pang2019fpinns,misyris2020physics,jagtap2020adaptive,cai2021physics,krishnapriyan2021characterizing,cai2022physics,cuomo2022scientific}. PINNs assume that a general, nonlinear PDE of the form
\begin{equation}\label{PINN1}
\dot{ \bs z} + \mathcal N [\bs z;\bs \mu]=\bs 0,
\end{equation}
is known to govern the physics taking place. Here, $\mathcal N$ is a nonlinear differential operator and $\bs \mu$ represents a set of parameters. Of course, $\bs z(\bs x,t)$, $\bs x\in \Omega$ represent the governing variables of the problem, defined in some open set $\Omega \subset \mathbb R^n$. The problem is then established so as to find, given some measurement on the system, the value of the variables governing it, $\bs z(\bs x,t)$, and the fittest parameters $\bs \mu$ that produce these measurements.

{  PINNs somehow resemble a collocation method. Given some measurements $\bs z^i(t_{\bs z}^i, \bs x_{\bs z}^i )$, the residual of Eq. (\ref{PINN1}) is defined as 
$$
\mathcal R = \dot{\bs z} + \mathcal N [\bs z;\bs \mu].
$$
By approximating $\bs z$ through a deep neural network, and applying automatic differentiation, one arrives at a neural network approximation to $\mathcal R$. By minimizing the mean squared error defined as
\begin{equation}\label{loss}
MSE = MSE_{\bs z} + MSE_{\mathcal R},
\end{equation}
with
$$
MSE_{\bs z} = \frac{1}{N_{\bs z}} \sum_{i=1}^{N_{\bs z}} \left| \bs z(t_{\bs z}^i, \bs x_{\bs z}^i )- \bs z^i \right|^2,
$$
and
$$
MSE_{\mathcal R} = \frac{1}{N_{\mathcal R}} \sum_{i=1}^{N_{\mathcal R}} \left|\mathcal R ( t_{\mathcal R}^i, \bs x_{\mathcal R}^i )\right|^2,
$$
with $\{t_{\bs z}^i, \bs x_{\bs z}^i, \bs z^i\}_{i=1}^{N_{\bs z}}$ denote measurements on the initial and boundary values and $\{ t_{\mathcal R}^i, \bs x_{\mathcal R}^i \}_{i=1}^{N_{\mathcal R}}$ represent the collocation points for $\mathcal R$.}

When learning the physics governing some given phenomenon, knowing in advance the PDE best describing it may seem as a too stringent condition, but the fact is that we already know many details about much of the physics surrounding us. In fact, in 1929, Paul Dirac said that \cite{dirac1929quantum},
\begin{quote}The underlying physical laws necessary for the mathematical theory of a large part of physics and the whole of chemistry are thus completely known, and the difficulty is only that the exact application of these laws leads to equations much too complicated to be soluble. \end{quote}

It seems therefore reasonable to think that some form of PDE could be envisageable for the phenomena under scrutiny. Even if this is not completely true---think again of an engineer trying to model a beam from observation: should he or she consider an Euler-Bernoulli-Navier model or a Timoshenko one?---PINNs constitute nowadays a very popular method for solving PDEs from data.

As discussed in Section \ref{Statistical} above, learning physics from data is even more difficult than that. The just presented approach does no guarantee, of course, that for any reason, the resulting prediction made by such a learned PDE will be consistent with the principles of thermodynamics that, as justified before, act as a restriction of a very high epistemic level.

If some symmetries of system are known in advance (remember that, thanks to the Noether's theorem, for each symmetry of the system there is a conserved quantity, see \cite{noether1918nachr} or its recent translation to English \cite{noether1971invariant}) these can be imposed beforehand. This is precisely the path followed in \cite{jin2020sympnets} for Hamiltonian systems of the form given by Eq. (\ref{Newton}). The PDE (\ref{PINN1}) is then restricted to an ordinary differential equation (ODE). We will come back to this approach in Section \ref{dynamical} below.

\subsection{Thermodynamics-based Artificial Neural Networks}

In a series of papers, I. Stefanou and coworkers presented recently a technique that employs thermodynamics as a restriction for learning constitutive laws \cite{masi2021thermodynamics,masi2020material,masi2021thermodynamics2,masi2022multiscale}. They coined the technique as Thermodynamics-based Artificial Neural Networks (TANNs). TANNs encode the two laws of thermodynamics by making use of automatic differentiation, thus ensuring by constructions the fulfillment of these laws, without the need to learn them from data. Thus, it also avoids the lack of fulfillment of these laws for previously unseen data.

Conservation of energy can be expressed as a PDE of the form
\begin{equation}\label{EC}
\rho \dot{e} = \bs \sigma \cdot \nabla^s \bs v - \nabla \cdot \bs q + \rho h,
\end{equation}
with $\rho$ the density, $\bs \sigma$ the stress tensor, $\bs v$ the velocity field, $\bs q$ the heat flux and $h$ the energy supply per unit mass. In turn, the second principle of thermodynamics can be expressed as
\begin{equation}\label{2PPO}
\rho (\theta \dot{s}-\dot{e}) + \bs \sigma \cdot \nabla^s \bs v -\frac{\bs q\cdot \nabla \theta}{\theta} \ge 0,
\end{equation}
with $s$ the specific entropy and $\theta$ the temperature.

TANNs try to avoid black-box ANNs by imposing the fulfillment of Eqs. (\ref{EC}) and (\ref{2PPO}). When fed with the current state of the material, $\bs z_t =\{ \bs \varepsilon_t, \Delta \bs \varepsilon, \bs \sigma_t, \theta_t, \zeta_t, \Delta t\}$, with $\bs \zeta$ a set of internal variables and $t$ the time, they produce an output composed by $\{\Delta \bs \zeta, \Delta \theta,F_{t+\Delta t},\Delta \bs \sigma\}$, where $F=E-S\theta$ is the Helmholtz free energy, $S=\rho s$ and $E$ is an energy potential (assumed to be rate-independent such that $\bs \sigma = \frac{\partial E}{\partial \bs \varepsilon}$ and $\theta = \frac{\partial E}{\partial S}$). 

TANNs actually learn from data the values of two scalars, the Helmholtz free energy $F$ and the dissipation $D= \rho (\theta \dot{s}-\dot{e}) + \bs \sigma \cdot \nabla^s \bs v$. For this to be possible, the increment in internal variables and temperature are also learned. The rest of the ingredients of Eqs. (\ref{EC}) and (\ref{2PPO}) are computed by automatic differentiation.

A somewhat related approach has recently been presented in \cite{patel2022thermodynamically}. An alternative approach based on exterior calculus is also developed in \cite{trask2022enforcing}.

\subsection{Variational Onsager Neural Networks}

{Huang and coworkers developed recently a technique based on the application of the Onsager variational principle \cite{onsager1931reciprocal,arroyo2018onsager}} which they coined as Variational Onsager Neural Networks, VONNs \cite{huang2022variational}. Again, their technique is based on the learning of two potentials: the free energy $F$ and the dissipation potential $D$. The resulting NN enforces strongly the fulfillment of the second law of thermodynamics.

For systems at constant temperature and free of inertial effects, the Onsager principle looks like
\begin{equation}\label{Onsager}
\min_{\bs \zeta} \mathcal R [\bs z, \bs \zeta],
\end{equation}
where $\mathcal R$ is the so-called Rayleighian and $\bs z$ and $\bs \zeta$ represent the state variables and the process (internal) variables, respectively. The Rayleighian takes the form
\begin{equation}\label{Ray}
\mathcal R [\bs z, \bs \zeta] = \dot{F} [\bs z, \bs \zeta] + D[\bs z, \bs \zeta] + P[\bs z, \bs \zeta],
\end{equation}
with $P[\bs z, \bs \zeta]$ the power supplied by external forces.

From the variational principle, Eq. (\ref{Onsager}), we arrive at
$$
\frac{\delta \dot{F}}{\delta \bs \zeta} + \frac{\delta D}{\delta \bs \zeta} + \frac{\delta P}{\delta \bs \zeta} =0.
$$

Both $F$ and $D$ are learned through a different NN, by minimizing the loss function {\it \`a la} PINN, see Eq.~(\ref{loss}). The convexity of the dissipation potential $D$ must be imposed explicitly, by making use of the input convex neural network paradigm \cite{amos2017input}. This may be also the case for the free energy potential $F$.

VONNs work after data for $\bs z$ and $\bs \zeta$. This means that, very much like TANNs, this technique needs for data on internal variables that are often impossible to measure. If these networks work is by the availability of synthetic data, coming from simulations. We will discuss further about this limitation in Section \ref{dynamical}. In any case, both techniques are able to impose in a soft manner the fulfillment of the second law of thermodynamics. 

In the next section we analyze a different family of techniques based on dynamical systems, ODEs, instead of PDEs, and discuss their relative advantages and disadvantages.

\section{Learning based on a dynamical systems analogy}\label{dynamical}

On a somewhat different setting, Weinan E showed that the learning problem has the same structure of a dynamical system \cite{E2017,WE2,barbaresco2021geometric}. In these works, it is suggested that supervised learning has the same structure of a dynamical system of the form
\begin{equation}\label{dyn}
\frac{d \bs z}{dt} = f(\bs z,t),\;\; \bs z(0)=\bs z_0,
\end{equation}
so that the flow map
$$
\bs z_0 \rightarrow \bs z(T ,\bs z_0),
$$
is produced by a non-linear function $f$ whose precise form is sought. This function can be found by employing classical regression methodologies such as linear regression, support vector machines \cite{hearst1998support}, or others. Of course, given the availability of the universal approximation theorem for neural networks, these appear as an appealing choice \cite{cybenko1989approximation,hornik1989multilayer,scarselli1998universal}. But the most interesting part of this approach is to recognize that, given the form of a dynamical system, all our previous knowledge in the field can be advantageously exploited to impose known structures in the system. The simplest structures one can think of are, of course, the Hamiltonian structure given by Eq. (\ref{Newton}) if the system is known to be conservative, or the gradient flow structure, for instance \cite{hohenberg1977theory}. For non-conservative variables, their evolution can be established after some potential $\mathcal R$ in the form \cite{CiCP-28-1639}
$$
\frac{d\bs z}{dt}= -\frac{\partial \mathcal R}{\partial \bs z}.
$$

As can be readily noticed, this framework assumes no known form for the problem at at hand. In sharp contrast to the framework of PINNs, whose governing PDE is assumed to be known, this dynamical systems equivalence assumes no previous knowledge on the physics taking place from which data are obtained. We explore this formalism more in detail in the following sections.

\subsection{Neural networks for conservative systems}

This framework has attracted the interest of many researchers in recent times. For instance, in \cite{jin2020sympnets} a method is developed for systems of the form given by Eq. (\ref{Newton}). To better explain this method, let us introduce some notation first.

A differentiable map $\phi : U\subset \mathbb R^{2N}\rightarrow \mathbb R ^{2N}$ is said to be symplectic if 
$$
\left( \frac{\partial \phi}{\partial \bs x} \right)^\top \bs L^\top \left( \frac{\partial \phi}{\partial \bs x} \right) = \bs L^\top.
$$
In particular, if $\phi_t(\bs z_0)$ is the flow map of a Hamiltonian system, it is a symplectic map,
$$
\left( \frac{\partial \phi_t}{\partial \bs z_0} \right)^\top \bs L^\top \left( \frac{\partial \phi_t}{\partial \bs z_0} \right) = \bs L^\top.
$$
Assume that we are in the position of obtaining experimental data about the system at hand in the form
$$
\mathcal D=\{ \bs z_i,\bs y_i=\phi_h(\bs z_i) \}_{i=1}^{\tt n_{\text{samples}}} ,
$$
with $h$ the time step size for an adequate numerical integration scheme on the problem. With these data we can think of constructing a feedforward neural network whose loss is of the form
$$
MSE = MSE_d+\lambda \cdot MSE_s,
$$
with
$$
MSE_d = \frac{1}{\tt n_{\text{samples}}} \sum_{i=1}^{\tt n_{\text{samples}}} \| \phi_h(\bs z_i) -\bs y_i \|^2
$$
the typical loss on the accuracy of the predictions and
\begin{equation}\label{loss2}
MSE_s = \frac{1}{\tt n_{\text{samples}}} \sum_{i=1}^{\tt n_{\text{samples}}} \left \| \left[ \left( \frac{\partial \phi_t}{\partial \bs z} \right)^\top \bs L^\top \left( \frac{\partial \phi_t}{\partial \bs z} \right)\right] (\bs z_i) -\bs L^\top   \right \|^2.
\end{equation}
$\lambda$ is a parameter to take into account the different relative size of both losses. This loss (\ref{loss2}) enforces the symplectic structure of the sought matrix $\bs L$, thus enforcing in turn the Hamiltonian character of the resulting approximation and its inherent conservative character. The resulting network works actually as a time integrator with time step $h$.

Many researchers have followed similar strategies. For instance, a year or so before Jin and coworkers, Greydanus et al. introduced the so-called Hamiltonian Neural Networks for canonical, discrete systems, in which the loss term is of the form \cite{greydanus2019hamiltonian}
\begin{equation}\label{loss3}
MSE = \left \| \frac{\partial \mathcal H}{\partial \bs p} - \frac{\partial \bs q}{\partial t} \right\|_2 + \left \| \frac{\partial \mathcal H}{\partial \bs q} + \frac{\partial \bs p}{\partial t} \right \|_2,
\end{equation}
where $\bs q$ and $\bs p$ represent, respectively, position and momenta of discrete particles. A similar approach is followed in \cite{PhysRevE.105.065305}. The same loss function in Eq. (\ref{loss3}) is employed in \cite{han2021adaptable}. However, in this work the authors solve parametric problems, so the input data set is collected for different values of these parameters, thus generalizing this type of networks, while keeping their conservative character. 

Even more sophisticate loss functions could be considered. For instance, Bertalan et al. assume that their data includes not only position and momenta, but also their derivatives. Their loss term is therefore of the form \cite{bertalan2019learning}
$$
\text{Loss} = \sum_{k=1}^4 c_kf_k,
$$
where each one of the four terms look like
$$
f_1 = \left \| \frac{\partial \mathcal H}{\partial \bs p} - \frac{\partial \bs q}{\partial t} \right\|_2,
$$
$$
f_2 = \left \| \frac{\partial \mathcal H}{\partial \bs q} + \frac{\partial \bs p}{\partial t} \right \|_2,
$$
$$
f_3 = (\mathcal H(\bs q_0, \bs p_0) - \mathcal H_0)^2,
$$
(this term serves for disambiguation only, since the Hamiltonian can be determined up to a constant. It is therefore assume to be known at a point $(\bs q_0, \bs p_0)$), and
$$
f_4 = \left \| \frac{\partial \mathcal H}{\partial \bs q}\dot{\bs q} + \frac{\partial \mathcal H}{\partial \bs p}\dot{\bs p} \right \|_2.
$$

The number of works that employs related approaches is huge and dates back to at least 1993 \cite{de1993class}. In \cite{david2021symplectic}, for instance, an improved training method is developed for this type of networks that is based on the symplectic character of the equations. Finzi et al. suggest a modification of the above techniques by working on cartesian coordinates and not in the phase space \cite{finzi2020simplifying}. To improve the expressiveness of these networks, Tong et al. introduce a method that adds Taylor series expansions designed with symmetric structure \cite{tong2021symplectic}. 

Always with the same Hamiltonian structure in mind, Chen et al. introduce them in the realm of recurrent neural networks, RNNs \cite{chen2019symplectic}. Even Generative Neural Networks exist under this Hamiltonian prism \cite{toth2019hamiltonian}. Again following a similar rationale, Di Pietro et al. improve the results by introducing a fourth-order time integration scheme in the learning scheme \cite{dipietro2020sparse}. Choudhary et al. focus their attention in the transition from ordered to chaotic systems, and show that Hamiltonian Neural Networks (HNNs) perform much better than classical, black-box NNs \cite{choudhary2020physics}. Other works also confirm the superiority of HNNs over classical NNs \cite{miller2020mastering,galimberti2021unified}.

But these architectures are interesting not only by their inductive biases, that force them to follow conservative dynamics. Galimberti and coworkers have also demonstrated that HNNs eliminate by construction the problem of vanishing gradients \cite{hochreiter1998vanishing}, present in many NN architectures \cite{galimberti2021hamiltonian}. Even a very recent survey paper has been written on this particular family of techniques \cite{chen2022learning}.

An alternative route to follow in this vast family of conservative phenomena is to employ Lagrangian, instead of Hamiltonian, formalisms. One advantage of doing so is the possibility of employing arbitrary coordinates instead of canonical coordinates. Although we will see later on that this does not constitute a true limitation for many systems of practical interest, canonical coordinates satisfy a set of rules in terms of the so-called Poisson bracket. The Lagrangian formalism assumes a data set composed by $\bs z = \{ \bs q, \dot{\bs q} \}_{i=1}^{\tt n_{\text{samples}}}$ (position and velocities of the particles). As it is well known, the Lagrangian formalism defines the so-called action functional as
$$
S= \int_{t_0}^{t_1} \left( T (\bs q, \dot{\bs q}) - V(\bs q) \right) dt,
$$
with $T$ the kinetic energy and $V$ the potential energy, so that a dynamical system will follow a path given by the minimum value of $S$. This value is obtained through the restriction to the so-called Euler-Lagrange equations,
$$
\frac{d}{dt} \frac{\partial \mathcal L}{\partial \dot{\bs q}} = \frac{\partial \mathcal L}{\partial {\bs q}} ,
$$
with $\mathcal L = T-V$ the Lagrangian of the system. This Lagrangian is precisely the objective to reconstruct from data. By standard algebraic manipulations, we arrive at
$$
\ddot{\bs q} = (\nabla_{\dot{\bs q}} \nabla_{\dot{\bs q}}^\top \mathcal L)^{-1} \left [ \nabla_{{\bs q}} \mathcal L - (\nabla_{{\bs q}} \nabla_{\dot{\bs q}}^\top \mathcal L)  \dot{\bs q}\right].
$$
From this expression it is therefore straighforward to define a loss function
$$
\text{Loss} = \| \ddot{\bs q}^{\mathcal L} - \ddot{\bs q}^{\text{true}} \|_2,
$$
so as to obtain an approximation to $\mathcal L$. This is the approach followed in \cite{cranmer2020lagrangian,roehrl2020modeling,allen2020lagnetvip,lee2002enhanced,zhong2020unsupervised,lutter2019deep,bhattoo2021lagrangian}, among others. Again, the amount of works devoted to this approach and their recent dates prove the interest of the community in these approaches.

\subsection{Neural networks for dissipative phenomena}

Despite the success in the development of network architectures that impose conservation of energy (achieved mainly in the last two years), many researchers have realized that, following the reasoning in Section \ref{Statistical}, dissipation is present in nearly every phenomenon of interest. Therefore, it is of utmost importance to develop techniques able to satisfy the principles of thermodynamics in the presence of dissipation. We make an overview of these techniques in this section.

\subsubsection{Deconstructing inductive biases}

Some authors, aware of the need to include dissipation in the formulation, have opted for its direct inclusion by just relaxing the fulfillment of the restrictions (inductive biases) introduced in the last section. This is the approach followed in the Symplectic ODE nets (symODEN) approach \cite{zhong2020dissipative,zhong2021benchmarking}, inspired by the literature on controlling dynamical systems. It is also the approach followed in a very recent approach \cite{gruver2022deconstructing}. Actually, both approaches are heavily influenced by the port-Hamiltonian approach to dynamical systems, which has a strong tradition in the introduction of dissipation and control in the formulation of dynamical systems  \cite{van2014port,rashad2020twenty,beattie2019robust}. Essentially, port-Hamiltonian systems consider an evolution of the system in the form
\begin{equation}\label{pH}
\begin{bmatrix}
\bs q \\ \bs p
\end{bmatrix} =
\left(
\begin{bmatrix}
\bs 0 & \bs I \\
-\bs I & \bs 0
\end{bmatrix}-\bs D(\bs q) 
\right)
\begin{bmatrix}
\frac{\partial \mathcal H}{\partial q} \\ \frac{\partial \mathcal H}{\partial p}
\end{bmatrix} + 
\begin{bmatrix}
\bs 0\\ \bs g(\bs q)
\end{bmatrix}\bs u,
\end{equation}
where, as can be noticed, dissipation is included through a symmetric, positive semi-definite matrix $\bs D$, and control in considered through an actuation $\bs u$. The formulation in Eq. (\ref{pH}) recovers the Hamiltonian structure if no dissipation nor control are present. However, as will be demonstrated later, this formulation does not guarantee the fulfillment of the principles of thermodynamics and can be considered, in some sense, phenomenological.

Of course, neural networks based upon this formulation have been developed in the last years \cite{massaroli2019port,cherifi2020overview,desai2021port,poli2020port,furieri2022distributed,eidnes2022port}. They all share the more or less the same ingredients and are based upon the formulation just discussed.

An alternative route is the one followed by Wang and coworkers \cite{wang2022approximately}. Given the equivalence of symmetries and conservation laws for different magnitudes, the authors choose to begin by employing equivariant neural networks \cite{cohen2016group,satorras2021n,keriven2019universal}. Then, these requirements on invariance are relaxed. But again, the lack of thermodynamic foundations of this method does not guarantee, in principle, the fulfillment of the principles of thermodynamics.

\subsubsection{Metriplectic neural networks}

As mentioned above, the port-Hamiltonian or relaxed equivariance formalisms do not guarantee but a phenomenological fulfillment of the laws of thermodynamics. There is no guarantee that, once faced to previously unseen data, or trained with noisy data, these networks will produce an output with the right amount of dissipation.

In order to develop a consistent formulation, let us assume that, at a given level of description, as discussed in Section \ref{Statistical}, the invariants of the system, $I(\bs z)$ can be expressed as a function of the resolved variables,
$$
I(\bs z) = \mathcal I(\bs z),
$$ 
for some suitable function $\mathcal I$. Let us also assume that the Hamiltonian of the system can be expressed as a function of these variables,
$$
\mathcal H (\bs z) = E(\bs z),
$$
with $E$ the actual energy of the system. In these circumstances, the Fokker-Planck equation (\ref{FP}) and, more particularly, its equivalent It\^o stochastic differential equation (\ref{Ito}) takes the form \cite{espanol1995statistical}
\begin{equation}\label{GENERIC1}
d\bs z = \left[ \bs L(\bs z) \frac{\partial E}{\partial \bs z} + \bs M(\bs z)\frac{\partial S}{\partial \bs z} + k_B\nabla \bs M(\bs z) \right]dt + d\tilde{\bs z},
\end{equation}
where $k_B$ is the Boltzmann constant, $\bs M(\bs z)$ is a symmetric, positive semi-definite dissipation matrix, $S$ is a second potential (the so-called Massieu potential, entropy at this level of description) and $d\tilde{\bs z}$ is a Wiener process that satisfies
$$
d\tilde{\bs z} = \bs B(\bs z)d\bs W(t),
$$
with $\bs B$ a non-square matrix satisfying
$$
\bs B(\bs z) \bs B(\bs z)^\top = 2k_B\bs M(\bs z).
$$
The importance of thermal fluctuations is controlled by the relative value of the Boltzmann constant $k_B$ with respecto to the average value of entropy. Given that $E$, $S$, $\bs L$ and $\bs M$ do not depend on $k_B$, if these effects are of low importance, we can take the limit $k_B\rightarrow 0$, resulting in
\begin{equation}\label{GENERIC2}
d\bs z = \bs L(\bs z) \frac{\partial E}{\partial \bs z} + \bs M(\bs z)\frac{\partial S}{\partial \bs z}.
\end{equation}
This same assumption, $k_B\rightarrow 0$, induces two additional consequences (we omit the proof, the interested reader can consult \cite{espanol1995statistical})
\begin{equation}\label{deg1}
\bs L(\bs z)\frac{\partial S}{\partial \bs z} = \bs 0,
\end{equation}
and
\begin{equation}\label{deg2}
\bs M(\bs z)\frac{\partial E}{\partial \bs z} = \bs 0,
\end{equation}
which constitute the ingredients of the celebrated General Equation for the non-Equilibrium Reversible-Irreversible Coupling, GENERIC, equations \cite{ottinger1997dynamics,ottinger2005beyond,grmela2018generic,grmela2019gradient,pavelka2018multiscale}. This type of formulations are also known as metriplectic formulations, since they combine metric and symplectic terms \cite{morrison1984bracket,morrison1986paradigm}. However, in GENERIC Eqs. (\ref{deg1}) and (\ref{deg2}), known as degeneracy conditions, play a fundamental role. They are key ingredients in the demonstration of the a priori satisfaction of the two laws of thermodynamics: conservation of energy in closed systems and non-negative entropy production. Indeed, given Eqs. (\ref{deg1}) and (\ref{deg2}), it is straightforward to prove that, given the anti-symmetry of $\bs L$,
$$
\dot{E}(\bs z) = \frac{\partial E}{\partial \bs z}\dot{\bs z} =0,
$$
and
$$
\dot{S} = \frac{\partial S}{\partial \bs z}\dot{\bs z} = \frac{\partial S}{\partial \bs z}\bs M(\bs z) \frac{\partial S}{\partial \bs z}\geq 0,
$$
given the positive semi-definiteness of $\bs M$. Therefore, the GENERIC structure appears to be much more interesting than port-Hamiltonian ones. It consistently guarantees the satisfaction of the laws of thermodynamics by construction. This makes GENERIC a very appealing choice for the construction of inductive biases in the learning of physical phenomena.

We assume that data sets $\mathcal{D}_i$ contain labelled pairs of a single-step state vector $\bs{z}_t$ and its evolution in time $\bs{z}_{t+1}$,
\begin{equation}
\mathcal{D}=\{\mathcal{D}_i\}_{i=1}^{N_{\text{sim}}},\quad\mathcal{D}_i =\{(\bs{z}_t,\bs{z}_{t+1})\}_{t=0}^{T},
\end{equation}
so that a neural network can be constructed by means of two loss terms, a data loss term that takes into account the correct prediction of the state vector time evolution using the GENERIC integrator, defined as
\begin{equation*}
\mathcal{L}^{\text{data}}_n=\left\Vert\frac{d\bs{z}^{\text{GT}}}{dt}-\frac{d\bs{z}^{\text{net}}}{dt}\right\Vert^2_2,
\end{equation*}
where $\Vert\cdot\Vert_2$ denotes the L2-norm. The choice of the time derivative instead of the state vector itself is to regularize the global loss function to a uniform order of magnitude with respect to the degeneracy terms.

A second loss term takes into account the fulfillment of the degeneracy equations,
\begin{equation*}
\mathcal{L}^{\text{deg}}_n=\left\Vert\bs{L}\dpar{S}{\bs{z}_n}\right\Vert^2_2+\left\Vert\bs{M}\dpar{E}{\bs{z}_n}\right\Vert^2_2.
\end{equation*}
This formulation gave rise to the so-called structure-preserving neural networks \cite{hernandez2021structure} and thermodynamics-informed neural networks \cite{hernandez2021deep,chinesta2020learning,hernandez2022thermodynamics}. These networks have been employed recently in the development of physcs perception with the help of computer vision techniques \cite{moya2021physics,moya2022physics}. 

The global loss term is a weighted mean of the two terms over the shuffled $N_{\text{batch}}$ batched snapshots.
\begin{equation}
\mathcal{L}=\frac{1}{N_{\text{batch}}}\sum_{n=0}^{N_{\text{batch}}}(\lambda \mathcal{L}^{\text{data}}_n+\mathcal{L}^{\text{deg}}_n).
\end{equation}
Note that the energy and entropy are learned through data about their gradients, so they are learnt up to an integration constant value. Note also that the activation functions must have a sufficient degree of continuity to allow for this.

Recently, alternative approaches have been developed to impose these degeneracy restrictions in hard form, instead of soft form. For instance, \cite{lee2021machine} employ a particular parametrization of the $\bs L$ and $\bs M$ matrices. Zhang and coworkers \cite{zhang2021gfinns} employ skew-symmetric matrices for forcing orthogonality. Notably, in this last work a universal approximation theorem is provided for this class of GENERIC-based networks, thus proving their expressivity.

An approach has been developed in which both inductive biases for the thermodynamic structure of the problem and a graph structure in the network are used in conjunction \cite{hernandez2022thermodynamics}. This approach has demonstrated to be very convenient for the development of learned simulators trained from finite element data. The result is a network for the problem of interest in which previously unseen geometric modifications can be introduced in the domain, as well as remeshings associated to these changes, without any decrease in accuracy.

Alternative versions based on GENERIC, but employing classical, piece-wise linear regressions instead of neural networks also exist, see \cite{gonzalez2019thermodynamically,gonzalez2019learning,moya2019learning,gonzalez2020data,moya2020physically}. More details on the geometric and thermodynamic structure of information can be found at \cite{barbaresco2020souriau}.

\begin{figure}[t]
	\centerline{\includegraphics[trim=210 50 200 0,clip,width=0.6\textwidth]{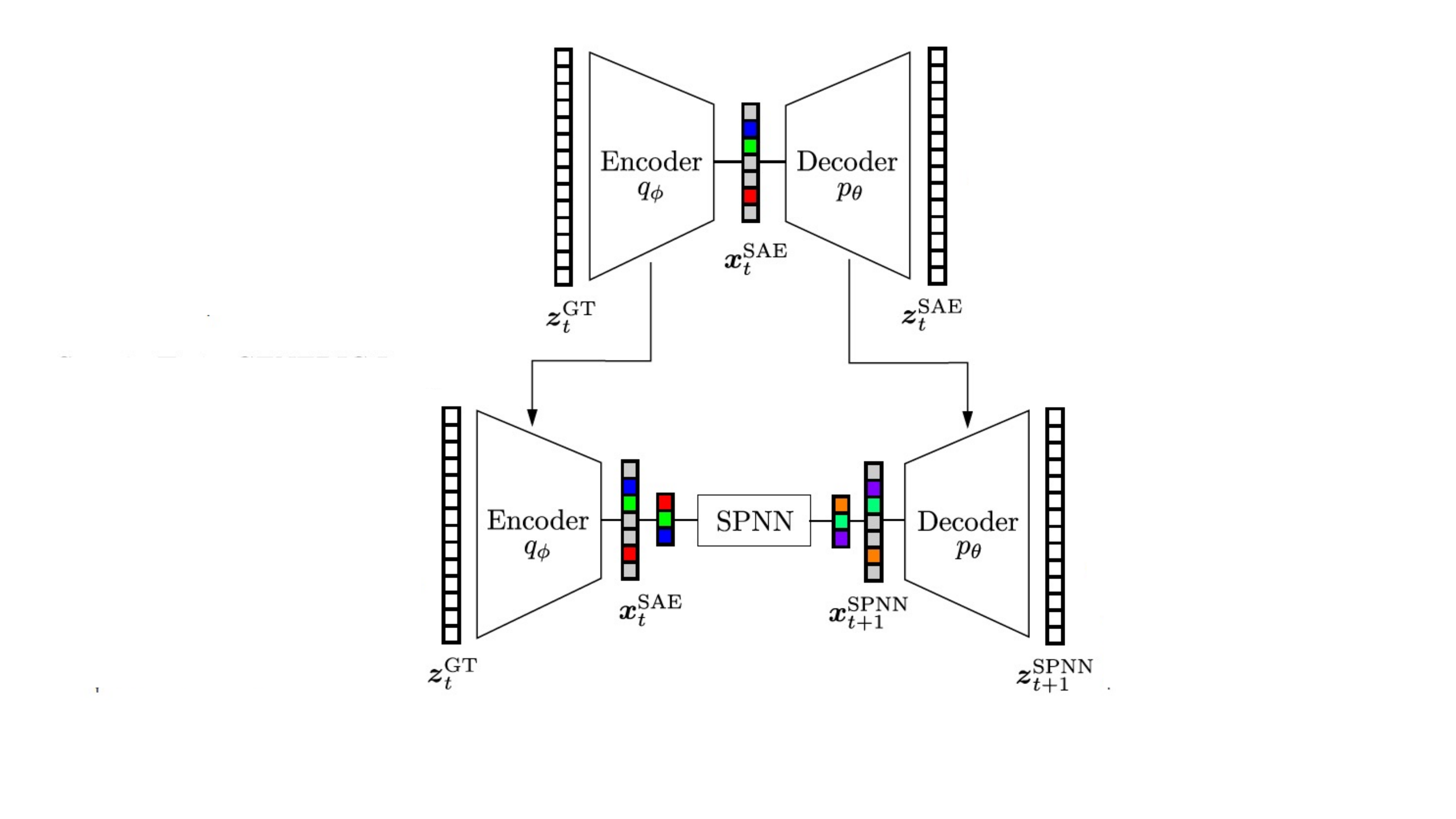}}
	\caption{Sketch of the architecture employed in \cite{hernandez2021deep} to learn reduced-order models of a given phenomenon.}
	\label{1}
\end{figure}

Something remarkable from this GENERIC approach is that is works whenever the set of state variables $\bs z$ is able to adequately represent the energy of the system \cite{espanol1995statistical}. This opens the possibility of constructing reduced-order models and apply TINNs to learn the evolution equations in this reduced-order manifold. This is the approach followed in \cite{hernandez2021deep}. In Fig. \ref{1} a sketch of this architecture is depicted. At a first instance, a sparse autoencoder is trained \cite{ng2011sparse}. This autoencoder is trained with the full-field variables $\bs z$. The autoencoder, that employs $L1$-norm enforcing sparsity, then finds a latent representation of the phenomenon $\bs x$. It is precisely in this latent space in which the method seeks for a GENERIC representation.

It is worth highlighting that the reduced-order modeling must follow this path: first reduce, then learn the GENERIC representation. The opposite (first learn the GENERIC model, then reduce) will make the ingredients of this formalism, $\bs L$, $\bs M$, $E$ and $S$ to loose their required properties.

\subsubsection{Generalized GENERIC structures}

One of the most frequent criticisms to the employ of GENERIC as an inductive bias when learning physical phenomena is that metriplectic formalisms are only able to consider quadratic potentials. However, this is not entirely true. While the early descriptions of metriplectic formalisms and of GENERIC itself included only quadratic dissipation potentials, see \cite{morrison1984bracket,morrison1986paradigm,ottinger1997dynamics,ottinger2005beyond}, this is no longer the case for modern descriptions of the GENERIC formalism \cite{grmela2019gradient,grmela2003framework}.

To develop such a theory, we first introduce the conjugate variables 
$$
\bs z^\ast = \frac{\partial S}{\partial \bs z}=S_{\bs z}.
$$
We also introduce the so-called dissipative thermodynamic forces $\mathcal X(\bs z^\ast)$. The third ingredient of the model is a dissipation potential
$$
\Xi = \Xi (\bs z, \mathcal X),
$$
with the following properties:
\begin{enumerate}
\item $\Xi$ is a real-valued and  regular function of $(\bs z, \mathcal X)$.
\item $\Xi (\bs z, 0)=0$.
\item $\Xi (\bs z, \mathcal X)$ reaches its minimum at $\mathcal X=\bs 0$.
\item $\Xi (\bs z, \mathcal X)$ is convex in a neighborhood of $\mathcal X=\bs 0$.
\item $\bs z^\ast\frac{\partial \Xi}{\partial \bs z^\ast} = K(\mathcal X, \frac{\partial \Xi}{\partial \mathcal X})>0$.
\end{enumerate}
With these conditions, we have that
\begin{equation}\label{GENERIC}
\dot{\bs z} = \bs L(\bs z) \frac{\partial E}{\partial \bs z} +\frac{\partial \Xi}{\partial \bs z^\ast}.
\end{equation}
Solutions to Eq. (\ref{GENERIC}) satisfy the conservation of energy and non-negative entropy production,
$$
\dot{S} = \frac{\partial S}{\partial \bs z} \frac{\partial \Xi}{\partial \frac{\partial S}{\partial \bs z }} = K\Big(\mathcal X, \frac{\partial \Xi}{\partial \mathcal X}\Big)>0.
$$

Eq. (\ref{GENERIC}) reduces to the standard GENERIC equation if $\mathcal X(\bs z^*) = \bs z^*$ and $\Xi (\bs z, \mathcal X) = \frac{1}{2} \bs z^* \bs M \bs z^*$, with $\bs M$ the usual dissipation matrix, symmetric and positive semi-definite.

Therefore, the elements in the GENERIC description of complex solids is composed by
\begin{enumerate}
\item the state variables $\bs z$.
\item the kinematics expressed by the Poisson bracket $\{a,b\}$,
\item the dissipative forces $\mathcal X$ and
\item three potentials, $E(\bs z)$, $S(\bs z)$ and $\Xi(\bs z,\mathcal X)$.
\end{enumerate}

To the best of the author's knowledge, there is no learning strategy developed on top of this last theory that, nevertheless, is worth exploring to extend the capabilities of metriplectic formalisms.

{\subsection{Strategies for open systems}

The variational Onsager strategy arising from Eq. (\ref{Ray}) presents one important advantage over strategies based upon metriplectic (GENERIC) formalisms: they are valid for externally-driven systems. On the contrary, GENERIC assumes by construction a closed system. Very recently, however, the metriplectic approach inherent in GENERIC has been extended to open systems in a port-Hamiltonian-like style \cite{hernandez2023port}. This new formalism can be seen as a sort of port-metriplectic approach in which dissipative, open systems can be tackled within the metriplectic framework by simply adding ports to the formulation, through which energy is exchanged with the environment. This opens the possibility to learn systems of systems, i.e., systems composed by sub-system, possibly of different nature, while ensuring the fulfillment of the right thermodynamic structure of the problem. Under the umbrella of the port-metriplectic formalism, the evolution of the open system can be analyzed as
\begin{multline}\label{PM}
    \dot{\bs z} = \lbrace \bs z,E\rbrace_{\text{bulk}} + [\bs z,S]_{\text{bulk}} \\
    = \lbrace \bs z,E\rbrace + [\bs z,S] - \lbrace \bs z,E\rbrace_{\text{boun}} - [\bs z,S]_{\text{boun}},
\end{multline}
where $\lbrace\cdot,\cdot\rbrace$ is the Poisson bracket and $[\cdot,\cdot]$ represents the dissipative bracket that constitute the conservative and dissipative, respectively, terms of GEN\-ERIC. In fact, the reader can notice that both brackets are decomposed additively into bulk and boundary contributions. The degeneracy conditions, Eqs. (\ref{deg1}) and (\ref{deg2}) are satisfied  by the bulk operators only.

}

\section{Conclusions}\label{conclusions}

We have explored the recent  interest in developing learning strategies for physical phenomena that take into account inductive biases originated from the non-equilibrium  thermodynamic theory. The growing interest in this field is motivated by the interest of the mechanics community on the development of strategies that avoid as much as possible black-box approaches to the problem. For our community, but also for industrialists, credible methods are necessary. These days somewhat resemble the early days of finite elements. By that times, industrialists did not understand how finite elements worked. This motivated a huge effort of research  within the applied maths and engineering communities that soon dissipated any doubt. If we want these data driven techniques to be adopted by the industry in a near future, we should provide responses, very much like the ones given for finite elements. This will make it necessary a joint effort from the applied maths community, from the engineering community but also from the physics and thermodynamics community, as we have tried to demonstrate in this paper.

In any case, so far these techniques have already demonstrated a substantial increase in robustness and credibility with respect to black-box approaches. It is also worth noting that something that we have already learnt is that the more physics knowledge we add to the learning process as an inductive bias, the less data will be necessary for the same level of accuracy and, conversely, the lower the error on the approximation will be. 

In general, we strongly believe that these techniques, and those that will be developed in the future, will change the way we think of simulation, paving the way for instantaneous responses for parametric problems to the designers and analysts.

\section*{Acknowledgement}

This work has been partially funded by the Spanish Ministry of Science and Innovation, AEI /10.13039/501100011033, through Grant number PID2020-113463RB-C31 and the Regional Government of Aragon and the European Social Fund, group T24-20R. The authors also acknowledge the support of ESI Group through the their respective chairs.

%\bibliography{Learning_GENERIC}
%\bibliographystyle{unsrt}

\end{document}